\documentclass[]{article}
\usepackage[letterpaper]{geometry}
\usepackage{mtsummit2015}
\usepackage{times}
\usepackage{url}
\usepackage{latexsym}
\usepackage{natbib}
\usepackage{layout}
\usepackage{graphicx}
\usepackage{multirow}
\usepackage{enumerate}
\usepackage{subfiles}
\usepackage{multirow}
\usepackage{graphicx}
\usepackage{subfiles}
\usepackage{authblk}

\def\code#1{\texttt{#1}}

\begin{document}

\title{\bf Tensor2Tensor for Neural Machine Translation}  

\author{{\bf Ashish Vaswani}}
\author{{\bf Samy Bengio}}
\author{{\bf Eugene Brevdo}}
\author{{\bf Francois Chollet}}
\author{{\bf Aidan N. Gomez}}
\author{{\bf Stephan Gouws}}
\author{{\bf Llion Jones}}
\author[1, 3]{{\bf \L{}ukasz Kaiser}}
\author[2]{{\bf Nal Kalchbrenner}}
\author{{\bf Niki Parmar}}
\author[1, 4]{{\bf Ryan Sepassi}}
\author{{\bf Noam Shazeer}}
\author{{\bf Jakob Uszkoreit}}

\affil[1]{Google Brain}
\affil[2]{DeepMind}
\affil[3]{Corresponding author: lukaszkaiser@google.com}
\affil[4]{Corresponding author: rsepassi@google.com}

\maketitle
\pagestyle{empty}

\begin{abstract}
  Tensor2Tensor is a library for deep learning models that is well-suited
  for neural machine translation and includes the reference implementation of
  the state-of-the-art Transformer model.
\end{abstract}

\section{Neural Machine Translation Background}

Machine translation using deep neural networks achieved great success with
sequence-to-sequence models \citep{sutskever14,bahdanau2014neural,cho2014learning}
that used recurrent neural networks (RNNs) with LSTM cells \citep{hochreiter1997}.
The basic sequence-to-sequence architecture is composed of an RNN encoder which reads
the source sentence one token at a time and transforms it into a fixed-sized state vector.
This is followed by an RNN decoder, which generates the target sentence,
one token at a time, from the state vector.

While a pure sequence-to-sequence recurrent neural network can already
obtain good translation results \citep{sutskever14,cho2014learning},
it suffers from the fact that the whole input sentence
needs to be encoded into a single fixed-size vector. This clearly
manifests itself in the degradation of translation quality
on longer sentences and was partially overcome in \cite{bahdanau2014neural}
by using a neural model of attention. 

Convolutional architectures have been used to obtain good results in word-level
neural machine translation starting from \cite{KalchbrennerB13} and later in
\cite{MengLWLJL15}. These early models used a standard RNN on top of
the convolution to generate the output, which creates a bottleneck and hurts performance.

Fully convolutional neural machine translation without this bottleneck
was first achieved in \cite{extendedngpu} and \cite{NalBytenet2017}.
The Extended Neural GPU model \citep{extendedngpu} used a recurrent stack
of gated convolutional layers, while the ByteNet model \citep{NalBytenet2017}
did away with recursion and used left-padded convolutions in the decoder.
This idea, introduced in WaveNet \citep{wavenet}, significantly improves
efficiency of the model. The same technique was improved
in a number of neural translation models recently, including \cite{JonasFaceNet2017}
and \cite{slicenet}.

\section{Self-Attention}

Instead of convolutions, one can use stacked
self-attention layers. This was introduced in the Transformer model 
\citep{transformer} and has significantly improved state-of-the-art
in machine translation and language modeling while also improving the speed of training. Research continues in applying the model in more domains and exploring the space of self-attention mechanisms. It is clear that self-attention is a powerful tool in general-purpose sequence modeling.

While RNNs represent sequence history in their hidden state, the Transformer has no such fixed-size bottleneck. Instead, each timestep has full direct access to the history through the dot-product attention mechanism. This has the effect of both enabling the model to learn more distant temporal relationships, as well as speeding up training because there is no need to wait for a hidden state to propagate across time. This comes at the cost of memory usage, as the attention mechanism scales with $t^2$, where $t$ is the length the sequence. Future work may reduce this scaling factor.

\begin{figure}
  \centering
  \includegraphics[scale=0.6]{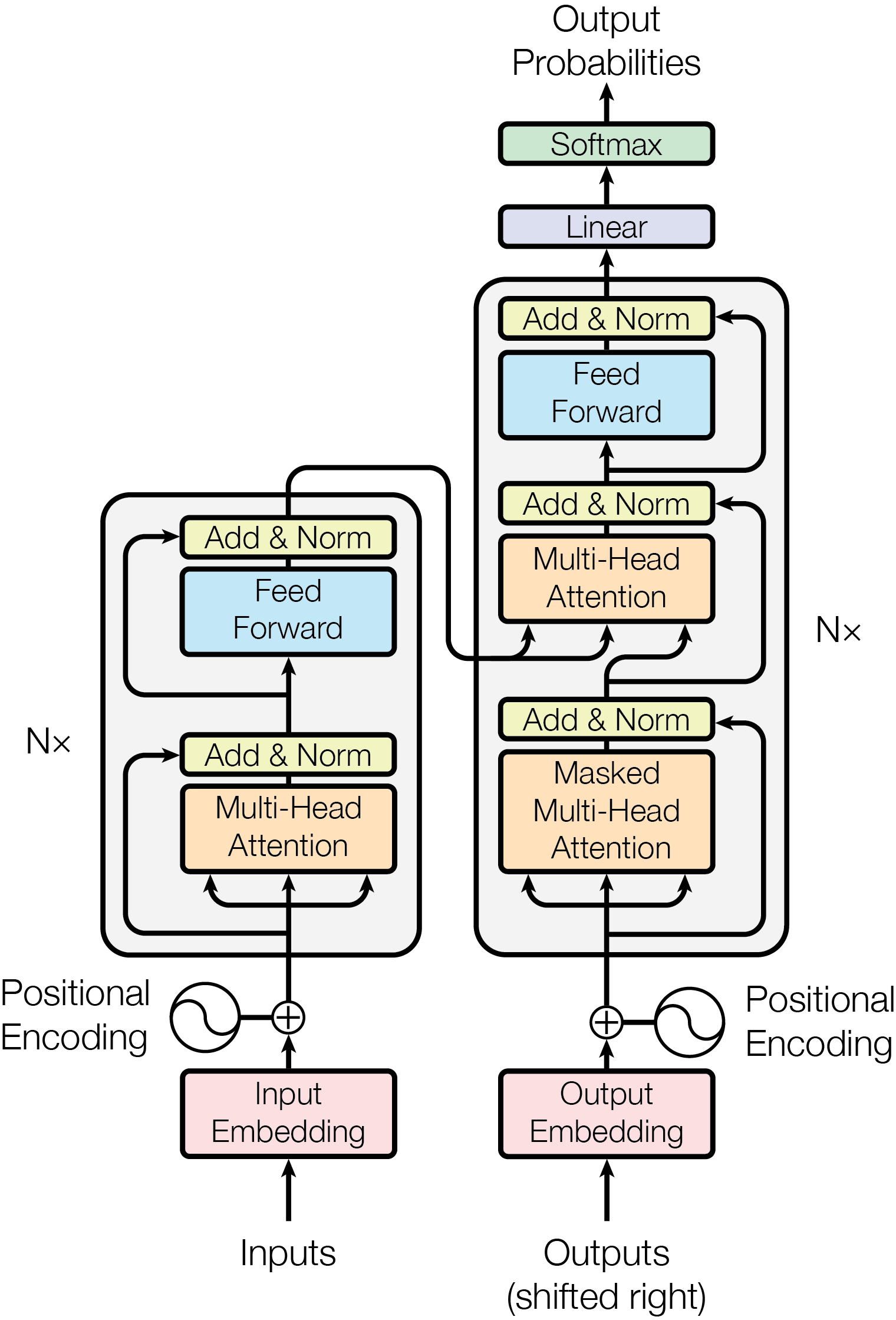}
  \caption{The Transformer model architecture.}
  \label{fig:transformer}
\end{figure}

The Transformer model is illustrated in Figure~\ref{fig:transformer}.
It uses stacked self-attention and point-wise, fully connected layers for both the encoder and decoder, shown in the left and right halves of Figure~\ref{fig:transformer} respectively.

Encoder: The encoder is composed of a stack of identical layers. Each layer has two sub-layers. The first is a multi-head self-attention mechanism, and the second is a simple, positionwise fully connected feed-forward network.

Decoder: The decoder is also composed of a stack of identical layers. In addition to the two sub-layers in each encoder layer, the decoder inserts a third sub-layer, which performs multi-head attention over the output of the encoder stack.

More details about multi-head attention and overall architecture can be found in \cite{transformer}.

\subsection{Computational Performance}

\begin{table}[t]
\caption{
  Maximum path lengths, per-layer complexity and minimum number of sequential operations for different layer types. $n$ is the sequence length, $d$ is the representation dimension, $k$ is the kernel size of convolutions and $r$ the size of the neighborhood in restricted self-attention.}
\label{tab:op_complexities}
\begin{center}
\vspace{-1mm}

\begin{tabular}{lccc}
\hline
Layer Type & Complexity per Layer & Sequential & Maximum Path Length  \\
           &             & Operations &   \\
\hline
\rule{0pt}{2.0ex}Self-Attention & $O(n^2 \cdot d)$ & $O(1)$ & $O(1)$ \\
Recurrent & $O(n \cdot d^2)$ & $O(n)$ & $O(n)$ \\

Convolutional & $O(k \cdot n \cdot d^2)$ & $O(1)$ & $O(log_k(n))$ \\
Self-Attention (restricted)& $O(r \cdot n \cdot d)$ & $O(1)$ & $O(n/r)$ \\

\hline
\end{tabular}
\end{center}
\end{table}

As noted in Table \ref{tab:op_complexities}, a self-attention layer connects all positions with a constant number of sequentially executed operations, whereas a recurrent layer requires $O(n)$ sequential operations.
In terms of computational complexity, self-attention layers are faster than recurrent layers when the sequence length $n$ is smaller than the representation dimensionality $d$, which is most often the case with sentence representations used by state-of-the-art models in machine translations, such as word-piece \citep{wu2016google} and byte-pair \citep{sennrich2015neural} representations.

A single convolutional layer with kernel width $k < n$ does not connect all pairs of input and output positions. Doing so requires a stack of $O(n/k)$ convolutional layers in the case of contiguous kernels, or $O(log_k(n))$ in the case of dilated convolutions \citep{NalBytenet2017}, increasing the length of the longest paths between any two positions in the network.
Convolutional layers are generally more expensive than recurrent layers, by a factor of $k$. Separable convolutions \citep{xception2016}, however, decrease the complexity considerably, to $O(k \cdot n \cdot d + n \cdot d^2)$. Even with $k=n$, however, the complexity of a separable convolution is equal to the combination of a self-attention layer and a point-wise feed-forward layer, the approach we take in our model.

Self-attention can also yield more interpretable models. In Tensor2Tensor, we can visualize attention distributions from our models for each individual layer and head. Observing them closely, we see that the models learn to perform different tasks, many appear to exhibit behavior related to the syntactic and semantic structure of the sentences.

\subsection{Machine Translation}
\begin{table}[t]
\begin{center}
\caption{The Transformer achieves better BLEU scores than previous state-of-the-art models on the English-to-German and English-to-French newstest2014 tests at a fraction of the training cost.  }
\label{tab:wmt-results}

\begin{tabular}{lccccc}
\hline
\multirow{3}{*}{Model} & \multicolumn{2}{c}{BLEU} & & \multicolumn{2}{c}{Training Cost} \\
& & & & \multicolumn{2}{c}{(in FLOPS * $10^{18}$)} \\
& EN-DE & EN-FR & & EN-DE & EN-FR \\ 
\hline
ByteNet \citep{NalBytenet2017} & 23.75 & & & &\\
Deep-Att + PosUnk \citep{DBLP:journals/corr/ZhouCWLX16} & & 39.2 & & & 100 \\
GNMT + RL \citep{wu2016google} & 24.6 & 39.92 & & 23  & 140\\
ConvS2S \citep{JonasFaceNet2017} & 25.16 & 40.46 & & 9.6 & 150\\
MoE \citep{shazeer2017outrageously} & 26.03 & 40.56 & & 20 & 120 \\
\hline
GNMT + RL Ensemble \citep{wu2016google} & 26.30 & 41.16 & & 180  & 1100\\
ConvS2S Ensemble \citep{JonasFaceNet2017} & 26.36 & \textbf{41.29} & & 77 & 1200\\
\rule{0pt}{2.2ex}Transformer (base model) & 27.3 & 38.1 & & \multicolumn{2}{c}{\boldmath$3.3$}\\
Transformer (big) & \textbf{28.4} & \textbf{41.8} & & \multicolumn{2}{c}{$23$} \\
\hline
\end{tabular}
\end{center}
\end{table}

We trained our models on the WMT Translation task.

On the WMT 2014 English-to-German translation task, the big transformer model (Transformer (big) in Table~\ref{tab:wmt-results}) outperforms the best previously reported models (including ensembles) by more than $2.0$ BLEU, establishing a new state-of-the-art BLEU score of $28.4$. Training took $3.5$ days on $8$ P100 GPUs.  Even our base model surpasses all previously published models and ensembles, at a fraction of the training cost of any of the competitive models.

On the WMT 2014 English-to-French translation task, our big model achieves a BLEU score of $41.8$, outperforming all of the previously published single models, at less than $1/4$ the training cost of the previous state-of-the-art model.

For the base models, we used a single model obtained by averaging the last 5 checkpoints, which were written at 10-minute intervals.  For the big models, we averaged the last 20 checkpoints. We used beam search with a beam size of $4$ and length penalty $\alpha=0.6$ \citep{wu2016google}.  These hyperparameters were chosen after experimentation on the development set.  We set the maximum output length during inference to input length + $50$, but terminate early when possible \citep{wu2016google}.

\section{Tensor2Tensor}

Tensor2Tensor (T2T) is a library of deep learning models and datasets designed to make deep learning research faster and more accessible.
T2T uses TensorFlow \citep{45381} throughout and there is a strong focus on performance as well as
usability. Through its use of TensorFlow and various T2T-specific abstractions, researchers
can train models on CPU, GPU (single or multiple), and TPU, locally and in the cloud, usually with no or minimal
device-specific code or configuration.

Development began focused on neural machine translation and so Tensor2Tensor includes many of the most successful NMT models and standard datasets. It has since added support for other task types as well across multiple media (text, images, video, audio). Both the number of models and datasets has grown significantly.

Usage is standardized across models and problems which makes it easy to try a new model on multiple problems or try multiple models on a single problem. See Example Usage (appendix \ref{usage}) to see some of the usability benefits of standardization of commands and unification of datasets, models, and training, evaluation, decoding procedures.

Development is done in the open on GitHub (http://github.com/tensorflow/tensor2tensor) with many contributors inside and outside Google.

\section{System Overview}

There are five key components that specify a training run in Tensor2Tensor:

\begin{enumerate}
    \item Datasets: The \code{Problem} class encapsulate everything about a particular dataset. A \code{Problem} can generate the dataset from scratch, usually downloading data from a public source, building a vocabulary, and writing encoded samples to disk. \code{Problem}s also produce input pipelines for training and evaluation as well as any necessary additional information per feature (for example, its type, vocabulary size, and an encoder able to convert samples to and from human and machine-readable representations).
    \item Device configuration: the type, number, and location of devices. TensorFlow and Tensor2Tensor currently support CPU, GPU, and TPU in single and multi-device configurations. Tensor2Tensor also supports both synchronous and asynchronous data-parallel training.
    \item Hyperparameters: parameters that control the instantiation of the model and training procedure (for example, the number of hidden layers or the optimizer's learning rate). These are specified in code and named so they can be easily shared and reproduced.
    \item Model: the model ties together the preceding components to instantiate the parameterized transformation from inputs to targets, compute the loss and evaluation metrics, and construct the optimization procedure.
    \item \code{Estimator} and \code{Experiment}: These classes that are part of TensorFlow handle instantiating the runtime, running the training loop, and executing basic support services like model checkpointing, logging, and alternation between training and evaluation.
\end{enumerate}

These abstractions enable users to focus their attention only on the component they're interested in experimenting with. Users that wish to try models on a new problem usually only have to define a new problem. Users that wish to create or modify models only have to create a model or edit hyperparameters. The other components remain untouched, out of the way, and available for use, all of which reduces mental load and allows users to more quickly iterate on their ideas at scale.

Appendix \ref{code_outline} contains an outline of the code and appendix \ref{usage} contains example usage.

\section{Library of research components}

Tensor2Tensor provides a vehicle for research ideas to be quickly tried out and shared. Components that prove to be very useful can be committed to more widely-used libraries like TensorFlow, which contains many standard layers, optimizers, and other higher-level components.

Tensor2Tensor supports library usage as well as script usage so that users can reuse specific components in their own model or system. For example, multiple researchers are continuing work on extensions and variations of the attention-based Transformer model and the availability of the attention building blocks enables that work.

Some examples:

\begin{itemize}
    \item The Image Transformer \citep{2018arXiv180205751P} extends the Transformer model to images. It relies heavily on many of the attention building blocks in Tensor2Tensor and adds many of its own.
    \item \code{tf.contrib.layers.rev\_block}, implementing a memory-efficient block of reversible layers as presented in \cite{DBLP:journals/corr/GomezRUG17}, was first implemented and exercised in Tensor2Tensor.
    \item The Adafactor optimizer (pending publication), which significantly reduces memory requirements for second-moment estimates, was developed within Tensor2Tensor and tried on various models and problems.
    \item \code{tf.contrib.data.bucket\_by\_sequence\_length} enables efficient processing of sequence inputs on GPUs in the new \code{tf.data.Dataset} input pipeline API. It was first implemented and exercised in Tensor2Tensor.
\end{itemize}

\section{Reproducibility and Continuing Development}

Continuing development on a machine learning codebase while maintaining the quality of models is a difficult task because of the expense and randomness of model training.
Freezing a codebase to maintain a certain configuration, or moving to an append-only process has enormous usability and development costs.

We attempt to mitigate the impact of ongoing development on historical reproducibility through 3 mechanisms:

\begin{enumerate}
    \item Named and versioned hyperparameter sets in code
    \item End-to-end regression tests that run on a regular basis for important model-problem pairs and verify that certain quality metrics are achieved.
    \item Setting random seeds on multiple levels (Python, numpy, and TensorFlow) to mitigate the effects of randomness (though this is effectively impossible to achieve in full in a multithreaded, distributed, floating-point system).
\end{enumerate}

If necessary, because the code is under version control on GitHub (http://github.com/tensorflow/tensor2tensor), we can always recover the exact code that produced certain experiment results.

\appendix
\section{Tensor2Tensor Code Outline} \label{code_outline}

\begin{itemize}
    \item Create \code{HParams}
    \item Create \code{RunConfig} specifying devices
    \begin{itemize}
        \item Create and include the \code{Parallelism} object in the \code{RunConfig} which enables data-parallel duplication of the model on multiple devices (for example, for multi-GPU synchronous training).
    \end{itemize}
    \item Create \code{Experiment}, including training and evaluation hooks which control support services like logging and checkpointing
    \item Create \code{Estimator} encapsulating the model function
    \begin{itemize}
        \item \code{T2TModel.estimator\_model\_fn}
        \begin{itemize}
            \item \code{model(features)}
            \begin{itemize}
                \item \code{model.bottom}: This uses feature type information from the \code{Problem} to transform the input features into a form consumable by the model body (for example, embedding integer token ids into a dense float space).
                \item \code{model.body}: The core of the model.
                \item \code{model.top}: Transforming the output of the model body into the target space using information from the \code{Problem}
                \item \code{model.loss}
            \end{itemize}
            \item When training: \code{model.optimize}
            \item When evaluating: \code{create\_evaluation\_metrics}
        \end{itemize}
    \end{itemize}
    \item Create input functions
    \begin{itemize}
        \item \code{Problem.input\_fn}: produce an input pipeline for a given mode. Uses TensorFlow's \code{tf.data.Dataset} API.
        \begin{itemize}
            \item \code{Problem.dataset} which creates a stream of individual examples
            \item Pad and batch the examples into a form ready for efficient processing
        \end{itemize}
    \end{itemize}
    \item Run the \code{Experiment}
    \begin{itemize}
        \item \code{estimator.train}
        \begin{itemize}
            \item \code{train\_op = model\_fn(input\_fn(mode=TRAIN))}
            \item Run the \code{train\_op} for the number of training steps specified
        \end{itemize}
        \item \code{estimator.evaluate}
        \begin{itemize}
            \item \code{metrics = model\_fn(input\_fn(mode=EVAL))}
            \item Accumulate the metrics across the number of evaluation steps specified
        \end{itemize}
    \end{itemize}
\end{itemize}

\section {Example Usage} \label{usage}

Tensor2Tensor usage is standardized across problems and models. Below you'll find a set of commands that generates a dataset, trains and evaluates a model, and produces decodes from that trained model. Experiments can typically be reproduced with the (problem, model, hyperparameter set) triple.

The following train the attention-based Transformer model on WMT data translating from English to German:

\begin{verbatim}
    pip install tensor2tensor

    PROBLEM=translate_ende_wmt32k
    MODEL=transformer
    HPARAMS=transformer_base
    
    # Generate data
    t2t-datagen \
      --problem=$PROBLEM \
      --data_dir=$DATA_DIR \
      --tmp_dir=$TMP_DIR
      
    # Train and evaluate
    t2t-trainer \
    --problems=$PROBLEM \
    --model=$MODEL \
    --hparams_set=$HPARAMS \
    --data_dir=$DATA_DIR \
    --output_dir=$OUTPUT_DIR \
    --train_steps=250000
    
    # Translate lines from a file
    t2t-decoder \
      --data_dir=$DATA_DIR \
      --problems=$PROBLEM \
      --model=$MODEL \
      --hparams_set=$HPARAMS \
      --output_dir=$OUTPUT_DIR \
      --decode_from_file=$DECODE_FILE \
      --decode_to_file=translation.en
\end{verbatim}

\end{document}